%% file: ijcai22-multiauthor.tex
\documentclass{article}
\usepackage{ijcai22}
\usepackage{tikz}    
\usepackage{tikz-qtree}
\usetikzlibrary{trees}
\usetikzlibrary{external}
\usetikzlibrary{shapes.geometric, arrows, shadows, calc,positioning,external}
\usepackage{times}

\usepackage{soul}
\usepackage{url}
\usepackage[hidelinks]{hyperref}
\usepackage[utf8]{inputenc}
\usepackage[small]{caption}
\usepackage{graphicx}
\usepackage{amsmath}
\usepackage{booktabs}
\usepackage{multirow}
\usepackage{latexsym}
\usepackage[normalem]{ulem}
\input{math_commands.tex}

\useunder{\uline}{\ul}{}
\newcommand{\CLS}{\mbox{CLS}}
\newcommand{\SEP}{\mbox{SEP}}
\newcommand{\EOT}{\mbox{EOT}}

\usepackage{subcaption}
\usepackage{tabularx}

\usepackage{color}

\title{Vision-and-Language Pretrained Models: A Survey}

\author{
Siqu Long$^1$
\and
Feiqi Cao$^1$\and
Soyeon Caren Han$^1$\And
Haiqin Yang$^{2}$\footnote{Corresponding author: hqyang@ieee.org.  Work done when Siqu and Feiqi were interned at IDEA.}\\
\affiliations
$^1$School of Computer Science, The University of Sydney, Australia\\
$^2$International Digital Economy Academy (IDEA), China\\
\emails
\{slon6753, fcao0492\}@uni.sydney.edu.au,
caren.han@sydney.edu.au,
hqyang@ieee.org
}

\begin{document}
\maketitle

\begin{abstract}
Pretrained models have produced great success in both Computer Vision (CV) and Natural Language Processing (NLP). This progress leads to learning joint representations of vision and language pretraining by feeding visual and linguistic contents into a multi-layer transformer, Visual-Language Pretrained Models (VLPMs). In this paper, we present an overview of the major advances achieved in VLPMs for producing joint representations of vision and language. As the preliminaries, we briefly describe the general task definition and genetic architecture of VLPMs. We first discuss the language and vision data encoding methods and then present the mainstream VLPM structure as the core content. We further summarise several essential pretraining and fine-tuning strategies. Finally, we highlight three future directions for both CV and NLP researchers to provide insightful guidance.
\end{abstract}

\section{Introduction}
In both Computer Vision (CV) and Natural Language Processing (NLP) communities, pretrained models have made significant progress. While CV researchers use VGG and ResNet using ImageNet to predict the categorical label of a given image, BERT~\cite{devlin2019bert} has been used and revolutionised many NLP tasks, such as natural language inference, and reading comprehension. Motivated by this, many cross-modal Vision-Language Pretrained Models (VLPMs) have been designed~\cite{vilbert,vlbert,uniter,oscar}. This pretrain-then-transfer learning approach to vision-language tasks naturally follows its widespread use in both CV and NLP. It has become the de facto standard due to the ease of use and solid representational power of large, publicly available models trained on large-scaled data sources. 

\begin{figure}[t]
    \centering
    \includegraphics[scale=0.40]{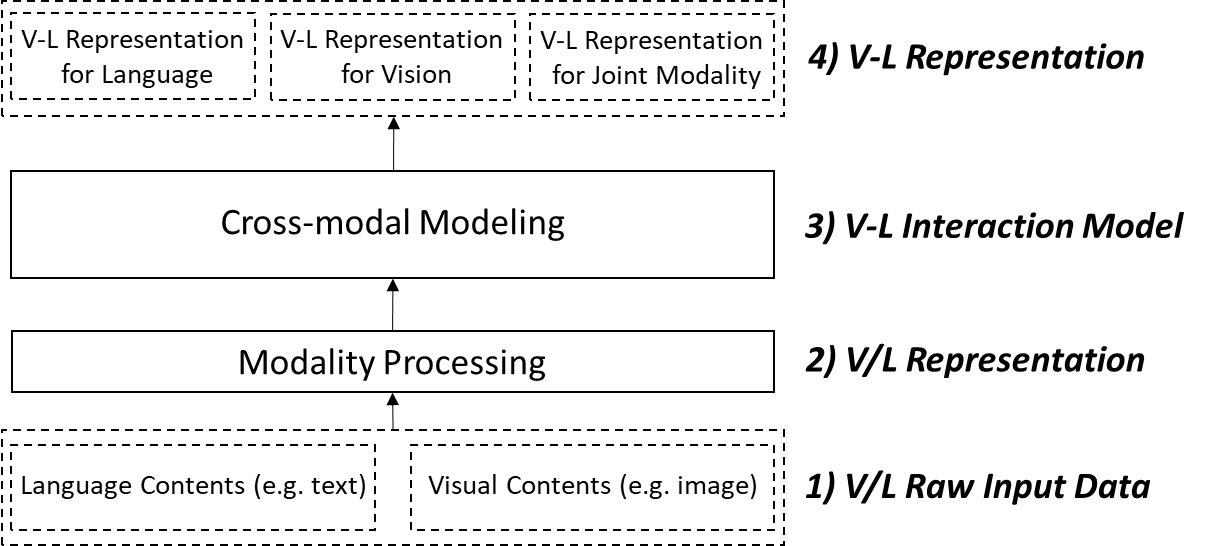}
    \caption{General architecture of VLPMs}
    \label{fig:vlp}\vspace{-5pt}
\end{figure}

In this paper, we present an overview of the rise and major advances achieved in the topic of VLPMs. Figure \ref{fig:vlp} illustrates a generic architecture of VLPMs. It involves the design of four main components: \textbf{1) V/L (Vision and Language) Raw Input Data} defines the representative \textit{raw data streams} from language and visual contents respectively, such as a single or multiple sentence(s) and one or a set of image(s). \textbf{2) V/L Representation} processes the raw data input into the desired format of \textit{modality representations} that can be used for \textbf{3) V-L (Vision-Language) Interaction Model}, which then enforces the cross-modal modeling between the two modalities. For instance, a common design is that the textual sentence is first tokenized and converted into the Bert-formatted input embedding while the image is processed into a set of spatial-aware RoI (Region of Interest) features. Those two modality representations are then concatenated and fed into the transformer encoder layers in which the cross-modal interaction is modeled via the multi-head self-attention mechanism. \textbf{4) V-L Representation} defines the possible cross-modal representations, which can be a \textit{V-L representation} for the single modality (i.e., Language or Vision) and/or the \textit{V-L representation} of joint modalities (i.e., Language and Vision). With the well-designed task supervision and learning guidelines from the pretraining, the \textit{V-L representation} finally learns to represent the generic cross-modal semantics, which would be transferred to help with the downstream V-L tasks via fine-tuning. This generic architecture applies to most of the existing VLPMs. The designs are various for each component, their pretraining strategies and transfer applications.

Existing surveys in this area have only partially reviewed some related tasks \cite{mogadala2021trends} or focused mainly on systematical analysis \cite{metaanalysis}. 
To the best of our knowledge, this is the first work that presents a comprehensive review of VLPMs. Our paper aims to provide both CV and NLP researchers insightful guidance for visual and language cross-modal learning via pretraining.

\section{Input Data Encoding}
\subsection{Language Encoding}\label{sec:l_stream}

Most VLPMs represent the \textit{language input} by taking a single textual sentence, which directly aligns with the image (visual modality) since their pretraining process mainly relies on pairwise image-text corpus \cite{visualbert,lxmert,vilbert,unicoder,uniter,ernie,oscar}. 
Some VLPMs use multiple sentences for representing the language input, especially with visual dialogue and multi-lingual settings \cite{vdbert,m3p,clcm}. 
Some special VLPMs apply visual input in the form of text and encode it as a part of the language input~\cite{oscar,tap}. For instance, OSCAR appends a set of object class tags detected from an image to the textual sentence as a language input to learn object-aligned V-L representation~\cite{oscar}. This early-fusion strategy of vision to language (V2L) at the raw input level serves as an anchor point to ease the cross-modal alignment learning.

To process the \textit{language input} into a suitable \textit{language representation} for cross-modal modeling, almost all VLPMs directly adapt the Bert-formatted input representation~\cite{devlin2019bert}, which sums up three types of learnable embeddings for each token in the textual sequence: 1) \emph{token embedding}, 2) \emph{position embedding}, and 3) \emph{segment embedding}. The \emph{token embedding} mostly follows the original Bert and encodes the textual sequence in the general form of ``$[\CLS]w_1w_2...w_i[\SEP]$" or with some minor modifications, where $w_i$ represents the $i$th tokenized (sub-)word based on the WordPiece vocabulary and $[\CLS]$/$[\SEP]$ are special tokens indicating the starting and ending of this sequence. Some models introduce additional special tokens to better align with specific data or pretraining tasks~\cite{vdbert,vlp,xgpt}, such as the $[\EOT]$ tokens for separating each dialogue turns \cite{vdbert}. The \emph{position embedding} is used exactly the same as in the original Bert, but the \emph{segment embedding} is adjusted to be \emph{modality embedding} in order to differentiate the two modalities or to further distinguish the multiple data streams within the language modality when multiple sentences are used, e.g., the segment tokens $A/B/C$ used in VL-Bert~\cite{vlbert}. As a special case, the \emph{visual feature} (See Sec.\ref{sec:v_stream}) can also be included into the language representation as the fourth type of embedding, which can be regarded as an early-fusion strategy of vision to language (V2L) at the representation level~\cite{vlbert,rvlbert}. With the Bert input representation format, it allows the VLPMs to enable initialising input embedding from Bert and directly adopt transformer encoder or its variants for the further cross-modal interaction modeling with a multi-head self-attention mechanism.

\noindent \textbf{Intra-modality Processing} In particular, some VLPMs apply \textit{additional intra-modality processing} with self-attention-based \textit{transformer} blocks to the aforementioned Bert-formatted language embeddings for further encoding the intra-modal contextual information, in order to better balance with the already high-level visual feature produced by the deep CNN-based extractor from the vision modality (see  Sec.~\ref{sec:v_stream}) and enable a robust single-modal representation of language~\cite{vilbert,lxmert,tden,tap,vldbert}. Instead, the resultant transformer output representation would be used as input for the V-L interaction model.

\subsection{Vision Encoding}\label{sec:v_stream}

With the processing of language input, the \textit{visual input} for VLPMs is normally a single image that directly aligns with the paired text input~\cite{visualbert,lxmert,vilbert,unicoder,uniter,ernie,oscar}, or a set of image(s) that are semantically correlated with each other~\cite{vldbert,prevalent}. For instance, the pretraining data for Vision-Language Navigation(VLN) is formed by a textual instruction with a group of panorama images along the navigation trajectory path~\cite{vldbert}.

Similar to the language representation, most studies encode \textit{visual input} into the Bert-style sequential representation, which consists of the aforementioned three major embeddings. The \emph{segment embedding} is created the same as that in language representation, but the \emph{token embedding} and \emph{position embedding} are modified to \emph{visual feature} and \emph{spatial position embedding} for capturing the visual semantics. Specifically, the \emph{visual feature} is extracted using the CNN-based feature extractors, the most common way in the CV domain. For this, the \textbf{\textit{granularity of representation}}, i.e., the grouping of image pixels into the sequential visual tokens, decides the alignment level of cross-modal modeling in the image content: \emph{1) RoI}-based VLPMs typically apply a pretrained Faster R-CNN object detector with a CNN backbone and extract the visual features of the detected object regions for the visual tokens~\cite{vilbert,lxmert,vdbert,uniter,unicoder,vlp} This is under the assumption that most image-text pairwise data is supposed to have its text describe the salient object (regions) in the corresponding image. Comparatively, some VLPMs simply split the whole image into continuous \emph{2) patches}~\cite{fashionbert,simvlm,cookie,soho} or even more fine-grained \emph{3) pixels}~\cite{pixelbert} as visual tokens for CNN-based feature extraction and refine the visual representation end-to-end, leading to significant speed improvement. More recent studies even propose the shallow, convolution-free embedding that utilizes simple linear projection to encode visual tokens at the \emph{patches}/\emph{pixels} level as in ViT~\cite{vit} for a further efficiency gain~\cite{vlmo,flava,vilt}. Besides, there are also VLPMs encoding each integral \emph{4) image} as a visual token by taking the pooling layer result from a CNN model (e.g., EfficientNet)~\cite{prevalent}.

Unlike the textual tokens with sequential positional relation, visual tokens entail \textbf{\textit{spatial positional relation}} instead, varying based on different granularity, which can be encoded by \emph{spatial position embedding}. RoI-based VLPMs commonly adopt the \emph{coordinate-based} position embedding\cite{vilbert,unicoder,vlp,uniter}, such as the 5-dimensional vector representing the normalized coordinates of the RoI bounding boxes and the fraction of image area. Comparatively, pixel/patch-based VLPMs represent the \emph{pixel location} using the 2D-aware vector for the row/column number \cite{soho}. PREVALENT \cite{prevalent} for Vision-Language Navigation (VLN) is the only image-based VLPM that specifies the spatial position embedding, which uses the elevation and heading angle relative to the agent to represent the positional relations among panorama images. The \emph{spatial position embedding} is normally combined with the \emph{visual feature} with or without \emph{segment embedding} via one or two FC layers followed by Layer Normalization to form the \textit{visual representation}.

\noindent \textbf{Intra-modality Processing} Since the transformer block used for additional intra-modality in language (See Sec.\ref{sec:l_stream}) is a global operator, whereas the CNN-based visual feature extractor is a local operator, it may lead to a different feature distribution between the two modalities~\cite{cookie}. Thus, some VLPMs also apply self-attention-based \textit{transformer} blocks to the initial visual representation (i.e., the Bert-style sequential representation described above) to align with the language intra-modality processing~\cite{lxmert,prevalent,tden,lightningdot,cookie}. Using transformers for intra-modality processing enables a \textit{modality-customized} encoding due to the freedom of selecting a different number of blocks for the two modalities. In practice, the language modality normally applies more transformer blocks than the vision modality for a better balance~\cite{lxmert,vilbert,prevalent}. Besides, several models adopt \textit{non-Transformer} processing, e.g., the AoANet~\cite{xgpt} and the Visual Dictionary-based mapping~\cite{soho}.

\section{V-L Interaction Model (V-LIM)}\label{sec:vl_arc}
There are two types of mainstream VLPM model structures: \textbf{(1) Single-stream (1-stream)} VLPMs~\cite{visualbert,uniter,oscar}, which directly fuse the \textit{initial language/visual representation} by using the joint cross-modal encoder at the initial stage, and \textbf{(2) Double-stream (2-stream)} VLPMs~\cite{lxmert,vilbert,visdialbert}, which separately apply the \textit{intra-modality processing} to two modalities along with a shared cross-modal encoder. However, this way of classification is mainly based on the perspective of intra-modality data-handling. In this section, we briefly review the model structure and the major output of VLPMs by focusing on their \textit{V-L interaction models (V-LIM)} instead, which can be: \textbf{1) Self-attention-based}, \textbf{2) Co-attention-based} or \textbf{3) VSE-based V-LIM}. 

\noindent \textbf{1) Self-attention-based V-LIM}\label{sec:sa-vlim}
Most single-stream~\cite{visualbert,unicoder,vlbert,pixelbert,fashionbert,uniter,oscar} and some of the double-stream VLPMs~\cite{soho} are considered as a \textit{self-attention-based V-LIMs} since they directly apply single-stream self-attention module to the modality representations for cross-modal modeling. They simply concatenate the language and visual representation (as introduced in Sec.\ref{sec:l_stream} and Sec.\ref{sec:v_stream}) to produce an integrated sequence representation so that a self-attention stream in the following transformer blocks can enforce the dense interaction modeling for both intra-modal (V-V\&L-L) and cross-modal (V-L). Following the transformer, the output representation of the global-level special token, such as $[\CLS]$, in the multi-modal sequence would be taken as the holistic \textit{joint V-L representation} for both pretraining and transfer learning. The output representation in the language or vision sequence could represent the contextualized V-L semantics and thus can be used as the \textit{V/L representation for Language or Vision}.

\noindent \textbf{2) Co-attention-based V-LIM}\label{sec:ca-vlim}
Comparatively, the VLPMs with \textit{co-attention-based V-LIM} decouple the intra- and cross-modal modeling processes. They keep two separate streams of transformer blocks for each modality that are entangled only at the specific \textit{cross-attention sub-layer(s)}. These sub-layers enforce exchange of the \textit{key} and \textit{value} in self-attention module between the two modality streams. In this way, they limit the cross-modal modeling only to those co-attention sub-layers while leaving the rest of the sub-layers independent to focus on intra-modality processing and modeling. This single-modality focus makes them align well with the definition of double-stream VLPMs~\cite{lxmert,vilbert,visdialbert,vldbert,prevalent,ernie}. As a result, the \textit{V/L representation for Language or Vision} can be taken from the output representation of the two separate modality streams whereas the \textit{V-L representation for joint modality} is derived by taking the global token representation from either modality sequence (or their aggregation).

\textbf{Encoder-decoder VLPMs} There are some VLPMs that apply a transformer-based encoder-decoder structure to empower the V-L generation ability. They can be categorized into either single-stream~\cite{vlp,vdbert,simvlm,unimo} or double-stream~\cite{xgpt,tden}. Regarding the V-LIM, few of them adopt \textit{unified encoder-decoder} structure with \textit{self-attention-based V-LIM}~\cite{vlp,vdbert,unimo}. Others try to emphasize the different peculiarities of the understanding/generation disciplines and instead use the conventional \textit{decoupled encoder-decoder} structure~\cite{simvlm,xgpt,tden}. Their V-LIM can be either \textit{self-} or \textit{co-attention-based} or even their combination, depending on how the attention modules in encoder/decoder are utilized for handling the multi-modalities. In general, these encoder-decoder VLPMs derive the \textit{V-L} representation via the cross-modal encoder in a similar way to those self-attention-based VLPMs.

\noindent \textbf{3) VSE-based V-LIM}
More recently, there is another emerging mainstream of VLPMs that simply utilizes the dual-encoder structured model with Visual-Semantic-Embedding(VSE)-based cross-modal contrastive learning~\cite{clip,align,lightningdot,cookie}. They tend to have a special focus on large-scale cross-modal retrieval with high demand for efficiency. They utilize the intra-modality processing to derive the vision and language representation (as described in Sec.\ref{sec:l_stream} and Sec.\ref{sec:v_stream}) between which the similarity-based cross-modal alignment would be modeled in the shared VSE space at the global level. This dual-encoder structure eradicates the fusion-style attention-based V-LIMs that are computation-costly and time-consuming. At the same time, it enables independent V/L representation encoding of both modalities, making the pre-computation possible for more efficient retrieval. The resultant \textit{V/L representation for language and vision} can be independently derived by the learned modality encoders whereas their fused representation can represent the \textit{V-L representation for joint modality}.

\section{Pretraining}\label{sec:pt}
In this section, we review the pretraining regarding the datasets, tasks and objective designs. 
\subsection{Datasets}\label{sec:pre_dataset}
Conceptual Captions (CC, roughly 3M) and SBU Captions (SBU, around 1M) of enormous size and diversified nature are the most commonly used webly collected datasets for VLPM pretraining~\cite{vilbert,vlp,tden,unicoder,clcm}. It is found that a larger sized corpus leads to better performance in downstream transfer tasks~\cite{vilbert,cookie,align}. Some simple Dual-encoder VLPMs~\cite{clip,align} collect even larger scaled corpus from the web, such as WIT (400M)~\cite{clip} and ALIGN(1.8B)~\cite{align}.

Another combination that leads to better domain adaptation is the \textit{out-of-domain+in-domain} datasets, where they define the web-based CC/SBU as \textit{out-of-domain} and the MS-COCO (COCO)/Visual Genome (VG) as the \textit{in-domain} datasets because most downstream tasks are built on them~\cite{uniter,ernie,lightningdot,vilt,vlmo,unimo}. The in-domain datasets can also be the \textit{task-specific} datasets that specifically come from the commonly evaluated downstream tasks (e.g., GQA/VQA2.0)~\cite{12in1,xgpt,vinvl,oscar,cookie}.

Some VLPMs target specialized tasks/domains such as Visual Dialogue (VD) and pretrain on only the \textit{domain/task-specific} dataset due to their reduced suitability to the aforementioned text-image datasets regarding domain nature or dataset format~\cite{fashionbert,kaleido,vdbert,prevalent}. There are also efforts of jointly pretraining with single-modal tasks on auxiliary \textit{single-modality} data source, i.e., non-paired collection of image or text, for reinforcing single-modal representations~\cite{vlbert,unimo,clcm,m3p,flava}.

\subsection{Tasks and Objectives}\label{sec:tasks_and_objectives}
There are three most commonly used \textit{cross-modal pretraining tasks}\footnote{The naming of these tasks/objectives may vary slightly in different papers.}: \textbf{1) Cross-modal Masked Language Modeling (CMLM)} extends the Masked Language Modeling (MLM) from Bert pretraining~\cite{devlin2019bert} to the multi-modal setting for learning the contextualized V-L representation, utilizing the bi-directional \textit{attention-based V-LIM}. This is proved to be helpful for transferring the pretrained language model Bert into multi-modal setting~\cite{vdbert} and thus makes it one of the most essential VLPM pretraining tasks~\cite{vlbert,pixelbert,uniter,m3p}. The task goal is to predict the masked tokens in text sequence based on the observation of their surrounding context, including both the unmasked textual tokens and all the visual tokens, by minimising the negative log-likelihood (NLL) loss. While most VLPMs mask out WordPiece sub-tokens as in Bert, some achieve better transfer performance through masking over semantically integral token groups such as complete word~\cite{vilt,fashionbert} or segment~\cite{ernie,unimo,oscar}.

\textbf{2) Cross-modal Masked Region Modeling (CMRM)} is a vision-supervised counterpart of CMLM, initially proposed by RoI-based VLPMs. It randomly masks out tokens in the visual sequence instead~\cite{uniter,lxmert,vilbert,unicoder,vlbert}. Thus far, there exist three variant objectives: \textbf{a) Region Label \textbf{C}lassification (CMRM\textsubscript{C})} predicts the object class of each masked region via minimizing the cross-entropy (CE) loss calculated based on the one-hot encoded object class from the object detector as ground-truth and the normalized distribution of the VLPM prediction over the possible object classes. To mitigate the potential classification error of the object detector, \textbf{b) Label \textbf{D}istribution Approximation (CMRM\textsubscript{D})} uses the probability distribution of the object class as a soft supervision by minimising the KL divergence loss between the object class distribution from the object detector and the normalized VLPM prediction distribution. Comparatively, \textbf{c) Region Feature \textbf{R}egression (CMRM\textsubscript{R})} learns to regress the VLPM output of each masked region into its input feature derived from the object detector using the L2 loss. It is commonly applied together with CMRM\textsubscript{C}~\cite{lxmert,m3p,unimo} or CMRM\textsubscript{D}~\cite{lightningdot,uniter} for more robust visual content modeling and joint cross-modal learning. Especially, when the visual token is represented at \textit{patch} level instead of RoI, CMRM can also be extended to masked \textit{patch} modeling~\cite{soho,kaleido,fashionbert}.

VLPMs with \textit{attention-based V-LIM} commonly treat \textbf{3) Cross-modal Alignment (CA)} as a binary classification problem, which aims to predict whether the input image-text pair is semantically matched or not based on the global V-L representation, by applying binary cross-entropy loss. The negative pairs can be sampled by randomly replacing either the image/text in the positive pair with another image/text from other data pairs in the corpus. On the other hand, dual-encoder based VLPMs with \textit{VSE-based V-LIM} all regard CA as a ranking problem for finding the best match and apply the contrastive learning objective for optimization~\cite{clip,lightningdot,align,cookie}. They learn the decoupled V/L representation in the semantically shared VSE space. Most of them adopt in-batch negatives for convenience, which simply treats the text/image from the remaining pairs within the batch as negatives. In addition to the two \textit{global-level alignment} CA objectives above, there are also several \textit{fine-grained alignment} variants~\cite{uniter,vilt} that try to enforce the matching between the fine-grained components such as visual regions/patches and textual words of the aligned image-text pairs. One essential difference between the two global-level and the fine-grained objectives is that the former introduces negative samples, which is found harmful for the fusion-based VLPMs with attention-based V-LIMs~\cite{vlbert}. Thus, some VLPMs~\cite{tden,flava,vlmo} introduce the single-modal encoders for its global-level CA while simultaneously keeping the fusion-based encoders for other tasks, to combine the advantages of the dual-encoder-based and fusion-based structures.

However, these three tasks cannot fulfil the goal of V-L generation tasks, such as Image Generation (IC)~\cite{vlp,xgpt,tden,simvlm}. Thus, encoder-decoder-based VLPMs especially design the pretraining tasks to involve a decoding process. Those unified encoder-decoder VLPMs simply extend the bidirectional CMLM to \textbf{seq2seq CMLM} as an auxiliary pretraining task for text generation~\cite{vlp,unimo,vdbert}. Other VLPMs with decoupled encoder-decoder emphasize the inherently different peculiarities of the understanding and generation disciplines, where the former entails the unrestricted information passing across modalities whereas the latter only involves visual-to-textual information passing. They keep the bi-directional cross-modal modeling only to the encoder while generating the text via a separate decoder in an auto-regressive manner~\cite{xgpt,tden,simvlm}.

From another perspective, applying downstream task such as the \textbf{IC} above for pretraining can be regarded as a kind of \textit{downstream-driven pretraining task}~\cite{lxmert,fashionbert}. Some special pretraining tasks especially designed for the specific downstream task/domain can also be downstream-driven~\cite{vdbert,visdialbert,prevalent,tap}, e.g., the Answer Prediction (\textbf{AnP}) for the Visual Dialogue(VD) task~\cite{vdbert}. Besides, there are VLPMs that jointly pretrain on additional text and/or image collection with the single-modal tasks~\cite{vlbert,vlmo,cookie,unimo}, including the single-modal multi-lingual settings~\cite{m3p,clcm}.

\section{Transfer Learning and Evaluation}\label{sec:tf}
In this section, we will introduce the downstream Visual-Linguistic understanding (V-L Understanding), Visual-Linguistic generation (V-L Generation) and single modal tasks that VLPMs mainly focus on.

V-L Understanding tasks refer to the tasks that require a model to capture both visual and linguistic semantics from the input and learn the correspondence between them. Most VLPMs apply a variety of V-L Understanding tasks for evaluation, while some just focus on a specific domain. Table~\ref{tab:t1} summarises the downstream tasks in V-L Understanding.

\noindent
\textbf{Visual Question Answering (VQA)} Most works~\cite{lxmert,vilbert,vlp,vlbert,villa,vlmo} formalize it as a classification problem to select an answer from a pool of common answers. However, SimVLM~\cite{simvlm} tries to decode open-ended answers in an auto-regressive manner to remove the limit posed by the fixed answer vocabulary. Among these models, Self-attention-based VLPMs work better than Co-attention-based VLPMs, and SimVLM outperforms all other VLPMs on the VQA v2.0 dataset.

\noindent
\textbf{Cross Modal Retrieval (CMR)} Some VLPMs~\cite{oscar,fashionbert} treat it as a binary classification task to predict whether each image-caption pair is matching, while others~\cite{uniter,tden,cookie,unicoder} solve it as a ranking problem to learn a similarity score to be maximized between positive pairs and minimized between negative pairs with a contrastive or CE loss.

\noindent
\textbf{Text Classification} Some VLPMs~\cite{lxmert,villa,12in1,pixelbert,uniter} work on the \textit{Natural Language for Visual Reasoning (NLVR)} task to perform binary classification based on special token representations, and some works~\cite{villa,12in1,uniter,unimo,soho,vilt} focus on the \textit{Visual Entailment (VE)} problem to evaluate their methods.

\noindent
\textbf{Visual Commonsense Reasoning (VCR)} The VCR task is usually divided into two sub-tasks to predict an answer and a rationale separately, and each sub-task is similar to answering a multiple-choice question out of four options. Therefore, most VLPMs~\cite{visualbert,vilbert,vlbert} formulate each sub-task as a binary classification problem for each option, or a 4-way classification problem as done by Unicoder-VL~\cite{unicoder}.

\noindent
\textbf{Referring Expression Comprehension (REC)} VLPMs~\cite{vilbert,vlbert,villa,12in1,uniter,ernie} working on this task usually perform a binary classification for each image region on whether it is the target of the phrase and select the region with the highest score at the inference stage. The VLPMs pretrained with in-domain datasets \cite{uniter,villa} generally work better on this task than those pretrained with only out-of-domain datasets.

\noindent
\textbf{Visual Relationship Detection (VRD)} Only RVL-BERT~\cite{rvlbert} focuses on this domain with two formulations. This problem can either be formulated as a ranking problem where the goal is to rank all possible subject-predicate-object triplets between any pair of object regions or as a binary classification problem to predict whether a given subject-predicate-object triplet holds.

\noindent
\textbf{Visual Dialogue (VD)} This problem can be formulated in a discriminative manner to select an answer from 100 candidates as done by VisDialBERT~\cite{visdialbert} and VDBERT~\cite{vdbert}, or in a generative manner to auto-regressively decode the answer as done by VDBERT~\cite{vdbert}.

\begin{table*}[t]
\centering
\small
\begin{tabularx}{\textwidth}{l|X}

\hline
\textbf{Tasks}                                & \textbf{Papers}  \\ \hline
VQA         &  ~\cite{visualbert},~\cite{lxmert},~\cite{vilbert},~\cite{vlp},~\cite{vlbert}, ~\cite{villa},~\cite{12in1},~\cite{pixelbert},~\cite{uniter},~\cite{oscar},~\cite{ernie},~\cite{tden},~\cite{vinvl},~\cite{soho},~\cite{kaleido},~\cite{simvlm},~\cite{unimo},~\cite{vlmo},~\cite{vilt},~\cite{flava}    \\ 
\hline
CMR         & ~\cite{vilbert}, ~\cite{unicoder}, ~\cite{villa}, ~\cite{12in1}, ~\cite{pixelbert}, ~\cite{fashionbert}, ~\cite{uniter}, ~\cite{oscar}, ~\cite{ernie}, ~\cite{tden}, ~\cite{vinvl}, ~\cite{m3p}, ~\cite{lightningdot}, ~\cite{soho}, ~\cite{unimo}, ~\cite{cookie}, ~\cite{clcm}, ~\cite{align}, ~\cite{vlmo}, ~\cite{vilt}, ~\cite{flava}        \\ 
\hline
NLVR & ~\cite{visualbert},~\cite{lxmert},~\cite{villa},~\cite{12in1},~\cite{pixelbert},~\cite{uniter},~\cite{oscar},~\cite{vinvl}, ~\cite{soho},~\cite{simvlm},~\cite{vlmo},~\cite{vilt}
                            \\ 
\hline
VE          &  ~\cite{villa}, ~\cite{12in1}, ~\cite{uniter}, ~\cite{unimo}, ~\cite{soho}, ~\cite{simvlm}, ~\cite{vilt}, ~\cite{flava}       \\ 
\hline
VCR         &  ~\cite{visualbert}, ~\cite{vilbert}, ~\cite{unicoder}, ~\cite{vlbert}, ~\cite{villa}, ~\cite{uniter}, ~\cite{ernie}, ~\cite{tden}       \\ 
\hline
REC         &  ~\cite{vilbert}, ~\cite{vlbert}, ~\cite{villa}, ~\cite{12in1}, ~\cite{uniter}, ~\cite{ernie}       \\ 
\hline
VRD         &  ~\cite{rvlbert}       \\ 
\hline
VD          &  ~\cite{vdbert}, ~\cite{visdialbert}       \\ 
\hline
VLN         &  ~\cite{vldbert}, ~\cite{prevalent}, ~\cite{vlnbert}     \\ 
\hline
\end{tabularx}
\caption{Categories of Downstream Tasks in Visual Language Understanding}\label{tab:t1}
\end{table*}

\noindent
\textbf{Visual Linguistic Navigation (VLN)} Some VLPMs~\cite{vldbert} try to solve this task in a discriminative setting by selecting the correct path among one positive path and several negative sample paths. In contrast, other VLPMs~\cite{vlnbert} try to solve it in a generative setting to make an action in between each state of the path by applying a reinforcement learning objective together with an imitation learning objective.

V-L Generation tasks generate texts/images when the other modality is included in the input. Most generative VLPMs~\cite{vlp,xgpt,oscar,vinvl,unimo,tden,simvlm,vivo} focus on \textbf{Image Captioning (IC)} task on COCO Captioning dataset by minimizing the NLL loss. Some of them~\cite{oscar,vinvl,simvlm,vivo} try the Nocaps task, which generates captions for images containing open domain novel objects. In addition, KaleidoBERT~\cite{kaleido} extends this task to the fashion domain, and TAP~\cite{tap} focuses on the TextCaps task (and the Text VQA task), which requires models to understand scene texts by applying CE loss at multiple decoding steps. VLPMs pretrained with extra in-domain datasets generally perform better except on the Nocaps. Similarly, multi-lingual VLPMs~\cite{simvlm} focus on the \textbf{Multi-modal Machine Translation (MMT)} task by minimizing the NLL loss. 

Despite focusing on multi-modal pretraining, some VLPMs also investigate how their models can generalize to single modal tasks. SimVLM~\cite{simvlm} and ALIGN~\cite{align} perform the image classification task of predicting object categories, and KaleidoBERT~\cite{kaleido} does it similarly to predict the category and subcategory of commercial products. Some VLPMs~\cite{simvlm,unimo} try to evaluate the natural language understanding capability of their models based on the GLUE benchmark tasks (or its subset). An improvement in linguistic tasks has been observed when pretrained with in-domain datasets, even with smaller pretraining data sizes.

\section{Conclusion and Future Research Directions}\label{sec:future}
This paper presents an overview of the recent advances in Vision-Language Pretrained Models for producing joint representations of visions and languages, especially for images and text pairs. We mainly summarise vision and language input encoding methods and mainstream VLPMs. We also discuss several useful pretraining and fine-tuning tasks and strategies. We do hope that the paper will provide insightful guidance for both CV and NLP researchers who work on joint and cross-modal learning.

In order to advance this field, there are several promising future directions for VLPMs.

\noindent\textbf{V-L Interaction Modeling.}
Although various VL interaction model extensions have been proposed in Section 3, there is still a significant challenge in aligning vision and language content. Most pretrained models focus on masking at the task level or input level, which does not directly align the features between image and text. Incorporating masking strategy at the embedding level has been shown to be effective~\cite{kaleido}. It is promising to investigate how to explicitly align the embedding features between image and text so that it can learn fine-grain representations. 

\noindent\textbf{VLPM Pretraining Strategy.} There still lacks systematic experiments and analysis on V-L-based multi-tasking synergy for VLPM pretraining. Few VLPMs explore multi-stage training~\cite{visualbert,vdbert,vlnbert,xgpt,uniter}, and 12-in-1~\cite{12in1} is the only one that tries to group the tasks of similar natures and test the performance boost from intra- and inter-group pretraining perspectives. It faces significant challenges to finding the answer in one step because it entails multiple factors such as selecting datasets (both multi-modal and single-modal), task design, task grouping, and order (multi-stage). Moreover, the effectiveness of the pretraining process may vary for different downstream targeted tasks. However, it is still worth exploring, step by step, how the V-L based multi-tasking can be implemented for VLPM pretraining that can generate the best transfer performance on the specific targeted domain/tasks, which will provide promising and valuable guidelines for the future development of VLPMs.

\noindent{\bf Training Evaluation.}  The trained VLPMs can be only evaluated during the downstream tasks.  This may waste a lot of computation cost when the models are defective.  It is worth exploring some metrics, such as perplexity, during the training procedure.  Hence, we can guarantee the performance of the trained VLPMs in advance. %in the training perspective.

\newpage

\bibliographystyle{named}
\bibliography{ijcai22}
\end{document}

%% file: math_commands.tex
\usepackage{amsmath,amsfonts,bm}
\usepackage{mathtools}
\def\eqref#1{equation~\ref{#1}}
\def\1{\bm{1}}

\DeclareMathAlphabet{\mathsfit}{\encodingdefault}{\sfdefault}{m}{sl}
\SetMathAlphabet{\mathsfit}{bold}{\encodingdefault}{\sfdefault}{bx}{n}